\documentclass[sigconf]{acmart}
\AtBeginDocument{%
  \providecommand\BibTeX{{%
    \normalfont B\kern-0.5em{\scshape i\kern-0.25em b}\kern-0.8em\TeX}}}

\copyrightyear{2022}
\acmYear{2022}
\setcopyright{acmcopyright}\acmConference[MM '22]{Proceedings of the 30th ACM International Conference on Multimedia}{October 10--14, 2022}{Lisboa, Portugal}
\acmBooktitle{Proceedings of the 30th ACM International Conference on Multimedia (MM '22), October 10--14, 2022, Lisboa, Portugal}
\acmPrice{15.00}
\acmDOI{10.1145/3503161.3547783}
\acmISBN{978-1-4503-9203-7/22/10}

\acmSubmissionID{200}

\usepackage{graphicx}
\usepackage{booktabs}
\newcommand\norm[1]{\left\lVert#1\right\rVert}
\usepackage{multirow}
\usepackage{bbding}
\usepackage{subfigure}
\usepackage{hyperref}

\begin{document}

\title{Dual Contrastive Learning for Spatio-temporal Representation}
\author{Shuangrui Ding}
\email{dsr1212@sjtu.edu.cn}
\affiliation{%
  \institution{Shanghai Jiao Tong University}
  \city{Shanghai}
  \country{China}
}
\author{Rui Qian}
\email{qr021@ie.cuhk.edu.hk}
\affiliation{%
  \institution{The Chinese University of Hong Kong}
  \city{Hong Kong}
  \country{China}
}

\author{Hongkai Xiong}
\authornote{Corresponding author.}
\email{xionghongkai@sjtu.edu.cn}
\affiliation{%
  \institution{Shanghai Jiao Tong University}
  \city{Shanghai}
  \country{China}
}

\renewcommand{\shortauthors}{Ding, Qian and Xiong.}

\begin{abstract}
Contrastive learning has shown promising potential in self-supervised spatio-temporal representation learning. Most works naively sample different clips to construct positive and negative pairs. However, we observe that this formulation inclines the model towards the background scene bias. The underlying reasons are twofold. First, the scene difference is usually more noticeable and easier to discriminate than the motion difference. Second, the clips sampled from the same video often share similar backgrounds but have distinct motions. 
Simply regarding them as positive pairs will draw the model to the static background rather than the motion pattern. 
To tackle this challenge, this paper presents a novel dual contrastive formulation.
Concretely, we decouple the input RGB video sequence into two complementary modes, static scene and dynamic motion. Then, the original RGB features are pulled closer to the static features and the aligned dynamic features, respectively. 
In this way, the static scene and the dynamic motion are simultaneously encoded into the compact RGB representation. We further conduct the feature space decoupling via activation maps to distill static- and dynamic-related features.
We term our method as \textbf{D}ual \textbf{C}ontrastive \textbf{L}earning for spatio-temporal \textbf{R}epresentation (DCLR). Extensive experiments demonstrate that DCLR learns effective spatio-temporal representations and obtains state-of-the-art or comparable performance on UCF-101, HMDB-51, and Diving-48 datasets.
\end{abstract}



\maketitle

\section{Introduction}
\label{intro}
Recently, self-supervised spatio-temporal representation learning has attracted great interest in the computer vision community. Compared with traditional supervised settings, the core of this learning scheme is to extract general representations from large-scale video data without resorting to human annotations. 
Instead of supervision from the costly manual labels, self-supervised learning obtains supervision from the unlabeled data themselves, enabling the utilization of millions of freely accessible videos on the Internet.

Inspired by the success in image domain~\cite{he2020momentum, chen2020simple}, contrastive-based methods have been expanded to spatio-temporal representation learning~\cite{qian2020spatiotemporal, feichtenhofer2021large} and achieved superior performance compared to previous pretext task-based methods~\cite{xu2019self,kim2019self,misra2016shuffle}.
\begin{figure}
    \centering
    \includegraphics[width=0.9\linewidth]{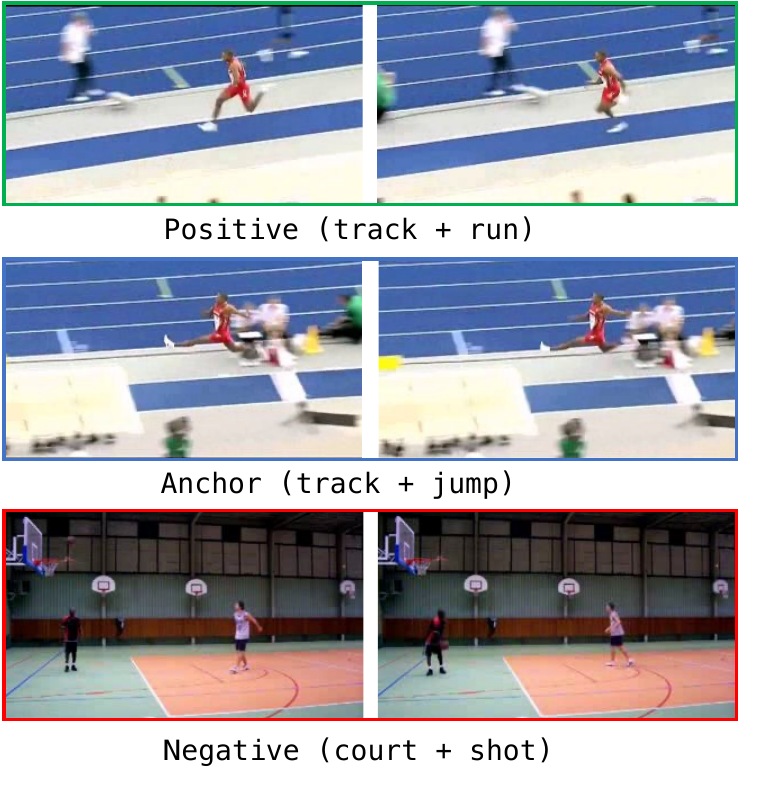}
    \caption{An illustration for positive and negative pair in spatio-temporal contrastive learning. 
    For positive pairs, the two clips have the same background (track) but distinct motions (run vs jump). And for the negative pair, the difference in the background scene (track vs court) is much more noticeable than the difference in motion (jump vs shot). These two phenomena cause static scene bias.}
    \label{fig:teaser}
\end{figure}
In particular, a common implementation of spatio-temporal contrastive learning is to sample two temporally different clips from each video, then regard pairs from the same (different) video as positive (negative) samples. The model is forced to draw `positive' pairs closer in the feature space and push apart the `negative' pairs. 
Under this formulation, the final representations are encouraged to capture the discriminative information, which greatly facilitates the performance.
However, previous works~\cite{wang2021removing, ding2022motion} reveal that this learning diagram tends to favor the static cues while focusing less on the motion. 
The potential reason lies in two aspects. 
First, the static scene bias exists in the positive pair formulation. In vanilla contrastive learning, we sample different temporal clips from each video. Those clips usually share a similar background but have a subtle difference in motion. 
As shown in Fig.~\ref{fig:teaser}, the positive pair owns the same background (track field) but different motions (jump vs run). If the model is encouraged to pull such biased positive pairs closer in the feature space, the model will naturally attend to the characteristics of the background scene but fail to capture the motion information.  
And for negative pairs, the background information still appears more salient than the motions. 
Take the negative pair in Fig.~\ref{fig:teaser} for instance, the scenes of the indoor basketball court and the outdoor track field almost dominate the entire screen. 
Thus, the considerable distinction between these two scenes seems to be sufficient for the model to push the negative pair away while the difference in motion patterns in the negative pair is hard to get noticed. 
To verify whether the background shortcut truly exists in contrastive learning, we utilize the static frame, which does not carry motion information, for analysis.
We regard the static frame as the positive sample of the original RGB clips to train the model. We find the model pretrained in this manner can achieve almost the same performance as the vanilla method (48.1\% vs 48.9\%). 
This empirical observation demonstrates that pulling RGB video pairs closer is basically equivalent to pulling the RGB video and the static frame closer, which reveals vanilla contrastive learning degrades the model to focus on background cues and learn static-biased representations.
More pieces of evidence are shown in Table~\ref{tab:mi}.

Hence, here comes a question on how to formulate the learning scheme that makes contrastive learning take both scene and motion into consideration. 
To delve into this problem, our paper proposes a novel approach named \textbf{D}ual \textbf{C}ontrastive \textbf{L}earning for spatio-temporal \textbf{R}epresentation (DCLR). 
We perceive scene and motion as two kinds of orthogonal and complementary information sources and decouple them at two levels.
We first decouple the static and dynamic information on the inputs and define a dual contrastive-based objective that enables the model to capture both static and dynamic features in video data. 
Particularly, given an RGB video sequence, we repeat a random frame along the temporal axis as the static data and regard the frame difference as the dynamic data.
We train the model by respectively minimizing two dual contrastive losses. One is the alignment between the original RGB sequence and static input, the other is between RGB and dynamic input. 
In addition, to emphasize motion learning, we mine the truly aligned motion positive pair of the RGB sequence across different videos. 
We do not view the clips from the same video as corresponding motion positive pairs. 
Rather, we maintain a dynamically updated feature extractor and retrieve the most similar motions as the corrected positive pairs.
Besides the static-dynamic decoupling in the input space, we further enhance the decoupling in the feature space. Specifically, we constrain that the RGB sequence feature activations should be consistent with the combination of static frame and frame difference activations. 
Meanwhile, we use the latter two activation maps to refine the static and dynamic related features in the RGB representation for dual contrastive learning. 
We evaluate the proposed DCLR on three action recognition downstream benchmarks, UCF-101, HMDB-51, and Diving-48, and manifest the effectiveness of each component in our framework. 
The experimental results demonstrate that DCLR enables contrastive spatio-temporal representation learning to resist the background shortcuts and achieve better generalization ability.

To sum up, our contributions are as follows: 
\begin{itemize}
    \item We formulate a novel self-supervised learning scheme, dual contrastive learning, motivated by the static bias in the spatio-temporal representation learning.
    \item We decouple static and dynamic cues in both data input and feature space to enhance dual contrastive learning.
    \item We achieve state-of-the-art or competitive results on downstream action recognition and video retrieval across UCF-101, HMDB-51, and Diving-48 datasets.
\end{itemize}

\section{Related Work}

\noindent\textbf{Self-supervised Representation Learning.}
The target of representation learning is to learn a transferable encoder that extracts desired characteristics from input data and filters redundant information. In traditional supervised learning like classification, \cite{he2016deep,xie2018rethinking,tran2018closer} directly use the category labels as the learning objective. While in self-supervised learning, it is nontrivial to define the objective. Early works design various pretext tasks, e.g., rotation~\cite{gidaris2018unsupervised}, colorization~\cite{kim2018learning}, jigsaw~\cite{misra2020self,doersch2015unsupervised}, to learn certain attributes. But the handcrafted pretext tasks are limited in performance. Later, contrastive learning promotes great progress in image representation learning~\cite{xu2021bag, oord2018representation,xu2021k}. It employs the consistency between multiple views of the same instance as self-supervisory signal~\cite{tian2020contrastive,chen2020simple}. Compared with the supervised settings, the multi-view constraint provides richer information, including semantics and some other instance-specific characters~\cite{zhao2020makes,xu2022seed}. Therefore, contrastive learning can obtain comprehensive representations that preserve unique information of each instance. 
However, in the video domain, due to the complex spatio-temporal structures, the naive multi-view constraint is far from satisfactory. In this work, we formulate a decoupled multi-view constraint to better fit the spatio-temporal nature.

\noindent\textbf{Spatio-temporal Representation Learning.}
Inspired by self-supervised learning in image domain, a line of works relies on pretext tasks to learn spatio-temporal representations. Since videos contain much richer characteristics, there are more pretext tasks: temporal ordering~\cite{misra2016shuffle,xu2019self,yao2020seco}, spatio-temporal puzzles~\cite{kim2019self,wang2020statistic}, playback speed prediction~\cite{jenni2020video,benaim2020speednet}, temporal cycle-consistency~\cite{wang2019learning,jabri2020space,li2019joint}, and future prediction~\cite{vondrick2016anticipating,villegas2017decomposing,luo2017unsupervised,behrmann2021unsupervised, han2019video,han2020memory}. 
Besides, some works expand the contrastive learning pipeline to the video domain by sampling different clips or modalities to formulate multi-view constraint~\cite{feichtenhofer2021large,qian2020spatiotemporal,wang2020self, han2020self, recasens2021broaden, alwassel2019self}. 
Recent works~\cite{wang2021removing, ding2022motion} observe that vanilla contrastive-based methods lead to static scene bias and attend less to temporal dynamic. 
To deal with it, our work reformulates the conventional contrastive learning as a dual learning problem, which encodes scene-debiased and motion-aware representations. \cite{zhang2022hierarchically, huang2021self} also adopt dual form of contrastive learning in video data. 
Specifically, \cite{zhang2022hierarchically} samples clips of the same timestamp for spatial contrast and samples clips of different timestamps for temporal contrast to separately learn spatial and temporal attributes. Since there is motion misalignment in temporally different clips, the static bias could still exist in~\cite{zhang2022hierarchically}. While ours can avoid this problem by directly maximizing the agreement between video clips with corresponding temporally aligned positive pairs. 
And for \cite{huang2021self}, they also conduct contrastive learning via three streams to extract RGB, static and dynamic features. But~\cite{huang2021self} only decouples representations at the level of data input. We extend this decoupling methodology to feature space. We adopt the attention map to refine the static-related and dynamic-related contrasts and restrict the complementarity of RGB features. The superior results compared to \cite{zhang2022hierarchically, huang2021self} manifest the effectiveness of our dual formulations.
\begin{figure*}
    \centering
    \includegraphics[width=0.9\linewidth]{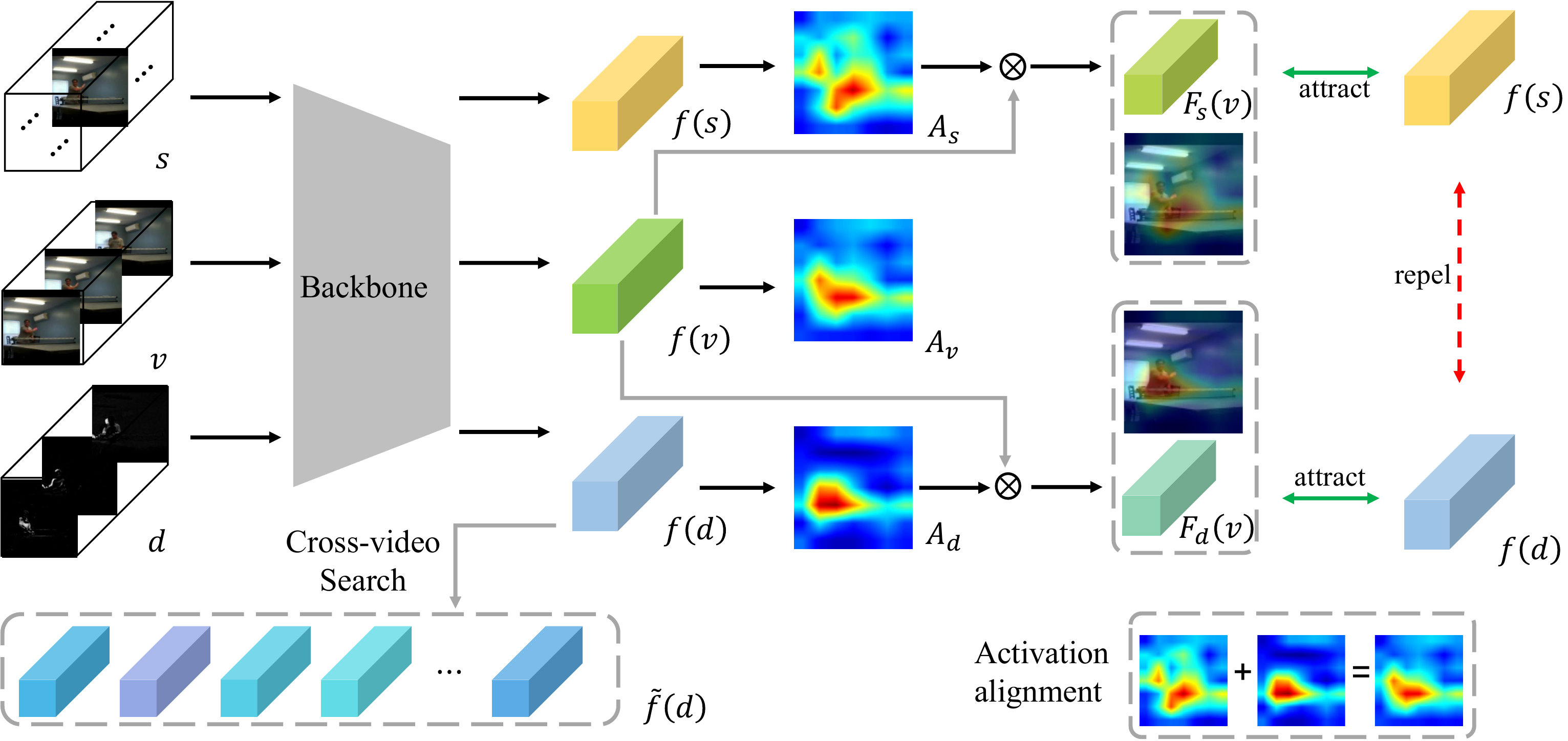}
    \caption{An overview of the proposed method. We first feed three data inputs $s,v,d$ into the backbone. We search the corresponding motion patterns in the large sample pool established by $\widetilde{f}(d)$ to correct the motion misalignment in the positive sample formulation. We utilize the activation maps as a concrete referrer to purify static and dynamic features, and employ the consistency constraint to let $f(v)$ cover the joint of $f(s)$ and $f(d)$. \textit{Best viewed in color}.}
    \label{fig:framework}
\end{figure*}

\noindent\textbf{Elimination of Background Bias in Video.}
The exploration of mitigating the background bias~\cite{he2016human, li2018resound, choi2019sdn, wang2018pulling} in the video emerges for a long time.
\cite{choi2019sdn} proposes to mitigate scene bias by augmenting the standard cross-entropy loss, where it leverages human-masked-out
videos to tell model whether the video contains action.
\cite{li2018resound} aims at alleviating static representation bias in existing datasets, where proposes a procedure to reassemble existing datasets. 
In self-supervised domain, other modality like optical flow~\cite{xiao2021modist, Li_2021_ICCV} are employed to emphasize motion information explicitly. 
To utilize the motion information implicitly in RGB, DSM~\cite{wang2021enhancing} and BE~\cite{wang2021removing} decouple the motion and context by deliberately constructing the motion-aware positive/negative samples through disturbance. 
Moreover, FAME~\cite{ding2022motion} proposes a copy-paste augmentation technique to keep the foreground motion intact.
It copies the foreground area onto the other background to resist the background shortcut. 
In this paper, we solve this bias problem from two perspectives. On the one hand, we decouple RGB into the easily accessible static frame and frame difference and train the model via dual contrastive loss. On the other hand, we sample corresponding motion patterns across videos in a large pool to weaken the bias in positive pair sampling.

\section{Approach}
In this section, we elaborate on our framework of dual contrastive learning. 
Our goal is to learn compact and rich feature representations from videos, which are not only scene-related but also contain temporal dynamic information. We first revisit vanilla spatio-temporal contrastive learning in Sec.~\ref{revisit}. And in Sec.~\ref{pair}, we illustrate the construction of positive (negative)
pairs in dual contrastive learning. 
Sec.~\ref{feature refinement} introduces the further decoupling in future space via activation maps. Finally, we give the full objective in Sec.~\ref{full}. 
\subsection{Spatio-temporal contrastive learning}
\label{revisit}
The vanilla spatio-temporal contrastive learning adopts instance discrimination~\cite{wu2018unsupervised} in a fully self-supervised manner. 
For ease of notation, we represent video data $v\in \mathbb{R}^{C \times T \times H \times W}$, where $C,T,H,W$ denote the dimensions of the channel, timespan, height, width, respectively. We notate the video encoder as $f:\mathbb{R}^{C \times T \times H \times W} \rightarrow \mathbb{R}^{D}$, where $D$ is the dimension of representation.
Similar to image domain~\cite{chen2020simple,he2020momentum,grill2020bootstrap}, it maximizes the similarity between two different views $v_i, v_j$ of one query sample $v$ and minimizes the similarity between negative pairs. 
Two different views of one video are sampled from two timestamps and then are processed by the temporal-consistent augmentation to reserve motion information~\cite{feichtenhofer2021large, qian2020spatiotemporal}. 
We simply adopt other samples from the mini-batch as negative sample.
Assuming the number of mini-batch is $N$, we take rest $2(N-1)$ examples in same mini-batch as negative samples. 
Having the positive pairs $(v_i, v_j)$, the loss function $\mathcal{L}_{VV}$ is formulated as
\begin{align}
    &\mathcal{L}_{VV} = I\left(f(v_i); f(v_j)\right) + I\left(f(v_j); f(v_i)\right), \\ 
    &I(f(v_i); f(v_j)) = -\log\frac{\exp(\text{sim}(f(v_i),f(v_j))/\tau)}{\sum_{k=1}^{2N} \mathbb{I}_{[k \neq i]} \exp(\text{sim}(f(v_i),f(v_k))/\tau)},
    \label{eq:infonce}
\end{align}
where $\mathbb{I}_{[k\neq i]} \in \{0, 1\}$ is an indicator function and $\tau$ is the temperature hyper-parameter. $\text{sim}(z_i,z_j)$ measures the cosine similarity between the latent representation, i.e., $\text{sim}(z_i,z_j)=z_i^T z_j/(\norm{z_i}_2\norm{z_j}_2)$.

As mentioned in the Sec.~\ref{intro}, this contrastive formulation possess severe background bias~\cite{wang2021removing, ding2022motion}. 
Two temporally different clips usually own similar static backgrounds but slightly diverse motions. 
In this way, contrastive learning would intuitively prioritize the background information alignment rather than motion. Such bias contributes to the weak generalization ability of the model. 
\subsection{Static-dynamic decoupling in data input}
\label{pair}
To mitigate the aforementioned background bias, we decouple the original input to perform dual contrastive learning, where we split the original video data into static frame and frame difference. 

\noindent\textbf{Static Frame.} We define static frame $s\in \mathbb{R}^{C \times T \times H \times W}$. Simply, we repeat a randomly selected frame to carry the static information without any dynamic motion. Mathematically,
\begin{equation}
    \label{static}
    s = [\underbrace{v_t,\cdots,v_t}_{T \text{ times}}] \quad \text{for any }t \in [1, T] 
\end{equation}
where $v \in \mathbb{R}^{C \times T \times H \times W}$ is the original clip and $t$ is the index of temporal dimension.

\noindent\textbf{Frame Difference.} We denote frame difference as $d\in \mathbb{R}^{C \times T \times H \times W}$. By differentiating adjacent frames iteratively, frame difference conveys natural motion information where moving areas possess a great magnitude and static cues are eliminated in this data input. Mathematically,
\begin{equation}
    \label{dynamic}
    d = v_{2:T+1} - v_{1:T},
\end{equation}
where $v \in \mathbb{R}^{C \times (T+1) \times H \times W}$ is the original clip and subscript is the index of temporal dimension.
Except for the frame difference, we also consider optical flow to convey motion information. 
Though the quality of optical flow looks more accurate, the extraction of the dense optical flow produces an unaffordable computational cost.  
Therefore, we adopt frame difference as a cheaper substitute.

Given complementary modalities $s$ and $d$, we can losslessly recover $v$ with simple union operation $s \cup d$. 
To this end, it is feasible to encode the feature of $v$ containing both characteristics of $s$ and $d$ at the same time.
Therefore, we transform the vanilla contrastive objective $\mathcal{L}_{VV}$ into a dual form as
\begin{align}
    &\mathcal{L}_{VV}\rightarrow \mathcal{L}_{VS}+\mathcal{L}_{VD},\\
    &\mathcal{L}_{VS}=I\left(f(v_i);f(s_j)\right)+I\left(f(v_j);f(s_i)\right),\\
    &\mathcal{L}_{VD}=I\left(f(v_i);f(d_j)\right)+I\left(f(v_j);f(d_i)\right),
\end{align}
where the function $I(\cdot;\cdot)$ is the same as Eq.~\ref{eq:infonce}.
$\mathcal{L}_{VS}$ ($\mathcal{L}_{VD}$) optimizes the alignment between video $f(v)$ and $f(s)$ ($f(d)$). 
By minimizing this dual loss term, $f(v)$ should contain both static and dynamic characters.

However, one concern in the decoupled contrastive optimization process is that $f(v)$ possibly collapses to the intersection of $f(s)$ and $f(d)$. 
In other words, $f(s)$ and $f(d)$ do not contain desired complementary information but only carry limited common information. 
This collapse exactly goes against our goal. 
Hence, in order to prevent the potential collapse problem, we enforce two decoupled representations orthogonal by maximizing a regularization term $\mathcal{L}_{SD}$, i.e.,
\begin{align}
\mathcal{L}_{SD}=
I\left(f(s_i);f(d_i)\right)+I\left(f(s_j);f(d_j)\right).    
\end{align}
Note that in contrast to $\mathcal{L}_{VS}$ and $\mathcal{L}_{VD}$ which take two different views of one video data, we use static frame and frame difference from the same view to ensure disparity between $f(s)$ and $f(d)$. 
The intuition is that if the model learns to attract positive pairs from the same view, it will easily attend to redundant information, which deviates from the vital characteristics. 
By maximizing $\mathcal{L}_{SD}$, we could guarantee the complementarity between $f(s)$ and $f(d)$, and let $f(v)$ simultaneously include much static as well as dynamic characteristics that help unbiased video understanding. We provide empirical results to support this formulation in the ablation study.

Now we have addressed the collapse problem by introducing the regularization term $\mathcal{L}_{SD}$, 
but the static bias in dual contrastive formulation remains. Concretely, the positive pair often shares similar backgrounds but differs in motions, i.e., given two views $v_i$ and $v_j$, the corresponding dynamic motions $d_i$ and $d_j$ do not always align. Hence, if we directly pull $f(v_i)$ and $f(d_j)$ closer through $\mathcal{L}_{VD}$, the model cannot learn helpful dynamic-related knowledge. 
Therefore, we conduct the cross-video search to figure out the truly aligned motion patterns as corrected positive motion pairs. 
Motivated by the queue mechanism in MoCo~\cite{he2020momentum}, we maintain a slowly changing frame difference feature extractor $\widetilde{f}$ which is updated every few epochs, and a queue of length $L$ to store extracted frame difference features.
In each iteration, given two-view frame difference input $d_i$ and $d_j$, we update the queue with $\widetilde{f}(d_j)$, and apply $\widetilde{f}(d_i)$ as query to retrieve similar pairs in the memory queue.

For a neat presentation, we omit the subscript and denote the query as $(v,d)$. We calculate the cosine similarity $S\in\mathbb{R}^{L}$ between $\widetilde{f}(d)$ and each frame difference feature in the queue. 
In this way, $S$ reveals the pair-wise similarity in dynamic motions. Hence, we can intuitively obtain cross-video motion pattern correspondence. 
We consider several variants of implementations in the retrieval stage. One simplest way is to retrieve the sample with highest similarity score, denoted as $(\widetilde{v},\widetilde{d})$. Then we use $(v,\widetilde{d})$ and $(\widetilde{v},d)$ to form modified $\mathcal{L}_{VD}$, i.e.,
\begin{align}
    \mathcal{L}_{VD} = I(f(v);f(\widetilde{d}))+I(f(\widetilde{v});f(d)).
    \label{joint}
\end{align}
Besides, we can select a subset of samples with topK highest similarity score  $[(\widetilde{v}_1,\widetilde{d}_1),(\widetilde{v}_2,\widetilde{d}_2),...,(\widetilde{v}_K,\widetilde{d}_K))]$, where $K$ is the number of samples in the selected subset. Meanwhile, we employ the cosine similarity score as prior knowledge. Then the loss calculation equals to  
\begin{align}
     \mathcal{L}_{VD} = \sum_{i=1}^{K} p_i [I(f(v);f(\widetilde{d_i}))+I(f(\widetilde{v_i});f(d))]
    \label{prob}
\end{align}
where $p_i = S[i]/\sum_{i=1}^{K}S[i]$ and we sort $S$ in descending order. Cross-instance retrieval is a common and effective way to align the high-level semantics shown in recent works~\cite{dwibedi2021little, koohpayegani2021mean}. 
In our work, the motivation for retrieving cross-video samples is different, where we aim to mitigate the motion pattern misalignment in positive pairs. 
Through this process, we fully leverage the natural characters of frame difference as well as the learned knowledge to mitigate the bias in the dual contrastive formulation. 
And we also explore reformulating $\mathcal{L}_{VS}$ in a similar manner. 
There is no significant gain in performance, which is concordant with our motivation that there only exists dynamic misalignment in positive pairs. 
The detailed discussions are displayed in the ablation study.

\subsection{Static-dynamic decoupling in feature space}
\label{feature refinement}
We further enhance static-dynamic decoupling in feature space. 
To do this, we consider an abstract measurement of the high-level features, which is the activation maps~\cite{baek2020psynet,zhou2016learning}. 
The activation maps involve richer information on the spatio-temporal distribution of the extracted features. Particularly, we obtain the class-agnostic activation maps~\cite{baek2020psynet} by calculating the summation of the feature maps $F(\cdot)\in \mathbb{R}^{C\times T \times H \times W}$ over the channel dimension.
For example, the activation $A_v \in \mathbb{R}^{T\times H\times W}$ for $F(v)$ is 
\begin{align}
    A_v[t,h,w] = \sum_{c=1}^C \left|F(v)\right|[c,t,h,w],
\end{align}
where $t,h,w$ denotes spatio-temporal index, $C$ is channel dimension. $A_s$ and $A_d$ are computed in the same fashion.

Naturally, $A_d$ attends more to the spatio-temporal areas that contain dynamic motions, while $A_s$ focuses on areas with discriminative static cues. 
Inspired by this, the activation $A_v$ for $f(v)$ should jointly highlight scene and motion related areas. Thus, we derive the activation alignment constraint:
\begin{align}
    \mathcal{L}_{ac} = \norm{A_v-(A_s+A_d)}_1.
\end{align}
We apply min-max normalization over the whole spatio-temporal dimensions to $A_v$ and $(A_s+A_d)$ in loss calculation. To stabilize the training, we only backpropagate the gradient of $\mathcal{L}_{ac}$ to $f(v)$ stream and stop the gradient to $f(s)$ or $f(d)$ stream. 
In this way, $A_v$ is encouraged to cover both dynamic and static reference areas and avoid falling into a trivial solution.

Besides the direct alignment in activation maps, we rely on $A_s$ and $A_d$ to purify the static and dynamic related features of $f(v)$. Particularly, we perform global weighted pooling on $F(v)$ to obtain the refined features.
Given $A_d$, we represent dynamic related features $f_d(v)$ as
\begin{align}
    f_d(v) &= \frac{\sum_{t,h,w}F(v)[t,h,w]\cdot A_d[t,h,w]}{\sum_{t,h,w}A_d[t,h,w]}.
\end{align}
Static related features $f_s(v)$ are obtained similarly. 
We now replace vanilla $f(v)$ with decoupled features in the dual formulation, i.e.,
\begin{align}
    I\left(f(v);f(s)\right)\rightarrow I\left(f_s(v);f(s)\right),
    \label{purify1}
    \\
    I\left(f(v);f(d)\right)\rightarrow I\left(f_d(v);f(d)\right).
    \label{purify2}
\end{align}
Through feature space decoupling, we filter out the possible noise that may interfere with our dual contrastive objective and enhance the representation ability.

\begin{table*}[!h]
\small
\centering
    \begin{tabular}{c|ccccc|cc}
        \hline
        Method & Backbone & Pretrain Dataset & Frames & Res. & Freeze & UCF-101 & HMDB-51 \\\hline
        CCL~\cite{kong2020cycle}& R3D-18 & Kinetics-400 & 16 & 112 & \Checkmark & 52.1 & 27.8 \\
        MemDPC~\cite{han2020memory} & R3D-34 & Kinetics-400 & 40 & 224 & \Checkmark & 54.1 & 30.5\\
        RSPNet~\cite{chen2021rspnet} & R3D & Kinetics-400 & 16 & 112 & \Checkmark & 61.8 & 42.8 \\
        MLRep~\cite{qian2021enhancing} & R3D & Kinetics-400 & 16 & 112 & \Checkmark & 63.2 & 33.4 \\
        FAME~\cite{ding2022motion} & R(2+1)D & Kinetics-400 & 16 & 112 & \Checkmark & 72.2 & 42.2 \\
        \textbf{DCLR(Ours)} & R(2+1)D & Kinetics-400 & 16 & 112 & \Checkmark & \textbf{72.3} & \textbf{46.4} \\
        \hline
        \hline
        VCP~\cite{luo2020video} & R3D & UCF-101 & 16 & 112 & \XSolidBrush & 66.3 & 32.2 \\
        IIC~\cite{tao2020self} & C3D & UCF-101 & 16 & 112 & \XSolidBrush & 72.7 & 36.8\\
        MLRep~\cite{qian2021enhancing} & R3D & UCF-101 & 16 & 112 & \XSolidBrush & 76.2 & 41.1 \\
        TempTrans~\cite{jenni2020video} & R(2+1)D & UCF-101 & 16 & 112 & \XSolidBrush & 81.6 & 46.4 \\
        \textbf{DCLR(Ours)} & R(2+1)D & UCF-101 & 16 & 112 & \XSolidBrush & \textbf{82.3} & \textbf{50.1} \\
        \hline
        3DRotNet~\cite{jing2018self} & R3D & Kinetics-400 & 16 & 112 & \XSolidBrush & 62.9 & 33.7 \\
        Pace Prediction~\cite{wang2020self} &R(2+1)D & Kinetics-400 & 16 & 112 & \XSolidBrush & 77.1 & 36.6 \\
        MemDPC~\cite{han2020memory}& R3D & Kinetics-400 & 40 & 224 & \XSolidBrush & 78.1 & 41.2 \\
        Pace~\cite{wang2020self} & R(2+1)D & Kinetics-400 & 16 & 112 & \XSolidBrush & 77.1 & 36.6
        \\
        VideoMoCo~\cite{pan2021videomoco} & R(2+1)D & Kinetics-400 & 32 & 112 & \XSolidBrush & 78.7 & 49.2
        \\
        MLRep~\cite{qian2021enhancing} & R3D & Kinetics-400 & 16 & 112 & \XSolidBrush & 79.1 & 47.6 \\
        TempTrans~\cite{jenni2020video} & R3D & Kinetics-400 & 16 & 112 & \XSolidBrush & 79.3 & 49.8 \\
        RSPNet~\cite{chen2021rspnet} & R(2+1)D & Kinetics-400 & 16 & 112 & \XSolidBrush & 81.1 &44.6 \\
        ASCNet~\cite{huang2021ascnet} & R3D & Kinetics-400 & 16 & 112 & \XSolidBrush & 80.5 & 52.3 \\
        SRTC~\cite{zhang2021incomplete} & R(2+1)D & Kinetics-400 & 16 & 112 & \XSolidBrush & 82.0 & 51.2 \\
        \textbf{DCLR(Ours)} & R(2+1)D & Kinetics-400 & 16 & 112 & \XSolidBrush & \textbf{83.3} & \textbf{52.7} \\
        \hline
    \end{tabular} 
    \caption{Results on action recognition downstream task. We present the backbone encoder, pretrain dataset, spatio-temporal resolution of each method. Freeze (tick) indicates \textit{linear probe}, and no freeze (cross) denotes \textit{end-to-end finetune}.}
    \label{tab:recognition}
    \vspace{-5mm}
\end{table*}

\subsection{The Full Objective}
\label{full}
The ultimate framework is illustrated in Fig.~\ref{fig:framework}. We first feed three data types $s,v,d$ into the video encoder $f$. It is worth noting that we adopt the same backbone for RGB, static frame, and frame difference since we empirically find that using separate backbones for three inputs achieves similar results with the same backbone. 
The underlying reason might be after the pre-processing normalization, the distribution of the three data types $s,v,d$ is not that different. 
Thus we adopt the same backbone to reduce parameters and training costs. 
For the dynamic motion branch, we maintain a dynamically updated feature extractor $\widetilde{f}$ to establish cross-video correspondence and correct the motion misalignment in the original positive pair formulation.
Then, we use activation maps $A_s$ and $A_d$, as soft masks to decouple static scene and dynamic motion related features for dual contrastive learning, and leverage the consistency between $A_v$ and $A_s + A_d$ to enhance the agreement between $f(v)$ and the union $f(s)\cup f(d)$.

Overall, the training loss consists of two parts, the dual contrastive term and the activation alignment term:
\begin{align}
    \mathcal{L} = (\mathcal{L}_{VS}+\mathcal{L}_{VD}-\mathcal{L}_{SD})+\lambda\mathcal{L}_{ac},
\end{align}
where $\lambda$ is the balancing hyper-parameter, set to 0.5 in default. Considering that the cross-video retrieval as well as the inferred $A_s$ and $A_d$ are not reliable in early training stage, we perform truly aligned motion positive pair correction in Eq.~\ref{joint} and feature refinement in Eq.~\ref{purify1} $\&$ \ref{purify2} after a few epochs. 

\section{Experiments}

\subsection{Implementation Details}

\noindent\textbf{Dataset.}
We use four popular video benchmarks for experiments, Kinetics-400~\cite{carreira2017quo}, UCF-101~\cite{soomro2012ucf101},  HMDB-51~\cite{kuehne2011hmdb} and Diving-48~\cite{li2018resound}. \textbf{Kinetics-400}~\cite{carreira2017quo} is a large-scale and high-quality dataset for action recognition, collected from realistic YouTube videos. Kinectis-400 contains over 240K video clips of 400 action classes. \textbf{UCF-101}~\cite{soomro2012ucf101} is an action recognition dataset consisting of over 13k clips covering 101 action classes.
\textbf{HMDB-51}~\cite{kuehne2011hmdb} is another action recognition dataset with 51 action categories and around 7,000 annotated clips.
\textbf{Diving-48}~\cite{li2018resound} involves 18k diving clips of 48 fine-grained diving categories, which majorly vary in motions and are no vast difference in the scenes. 
We pretrain the model on the training set of UCF-101 or Kinetics-400, and evaluate on the split 1 of UCF-101 and HMDB-51, or the V2 test set of Diving-48.

\noindent\textbf{Self-supervised Pretraining.}
We use R(2+1)D-18~\cite{tran2018closer}, with 14.4M parameters, as the video encoder, and share the same network to extract RGB, static frame, and frame difference features.
We randomly sample two temporally different clips in each video as two views and apply temporally consistent random resized crop and random horizontal flip to obtain the frame sequence. 
We decouple the static frame input $s$ and the frame difference input $d$ as described in Eq.~\eqref{static} $\&$ \eqref{dynamic}.
We apply color jitter and Gaussian blur to augment the RGB input $v$, static frame input $s$. 
In this way, $v$, $s$ and $d$ are all of spatio-temporal resolution $16\times 112\times 112$ and input to the same encoder.
We pretrain the model for 200 epochs on UCF-101 or 100 epochs on Kinetics-400. An SGD optimizer is adopted with the initial learning rate of $10^{-2}$ and weight decay of $10^{-4}$.
For the cross-video search, on UCF-101, we update the slowly changed feature extractor every 10 epochs with a queue length of 2048. On Kinetics-400, we update the extractor every 5 epochs with a queue length of 16384. 


\noindent\textbf{Action Recognition.} We use the pretrained parameters except for the last fully-connected layer for initialization. 
We employ two popular protocols to validate the self-supervised representations: (1) \textit{Finetune} the whole network with action labels; (2) Freeze the backbone and train the last linear classifier, denoted as \textit{linear probe}. During the inference phase, we follow the prevalent evaluation protocols~\cite{xu2019self, wang2020self} to uniformly sample ten 16-frame clips from each video, then center crop and resize them to $112\times 112$. We average the prediction of each clip as video-level action prediction and report Top-1 accuracy to measure the action recognition performance.

\noindent\textbf{Video Retrieval.} We extract spatio-temporal features from the pretrained model without any further training. Based on the cosine similarity, videos in the test set retrieve the Top-$k$ nearest neighbors in the training set.
If the category of the test set exists in the $k$ nearest neighbors, it counts as a hit.
Following~\cite{xu2019self, wang2020self}, we average representations of ten uniformly sampled clips. 
We report Top-$k$ recall R@k for evaluation.

\subsection{Evaluation on Downstream Tasks}

\begin{table}
    \small
    \centering
    \begin{tabular}{c|cc|c}
         \hline Method & Pretrain Dataset & Res. & Top-1 \\
         \hline
         Random Init. & - & - & 50.7 \\
         BE~\cite{wang2021removing} & UCF-101 & 224 & 58.8 \\
         FAME~\cite{ding2022motion} & UCF-101 & 224 & 67.8\\
         \textbf{DCLR(Ours)} & UCF-101 & 112 & 72.7 \\\hline
         BE~\cite{wang2021removing} & Kinectics-400 & 224 & 62.4 \\
         FAME~\cite{ding2022motion} & Kinectics-400 & 224 & 72.9\\
         \textbf{DCLR(Ours)} & Kinectics-400 & 112 & 75.1 \\
        \hline
    \end{tabular}
    \caption{Top-1 accuracy on Diving-48 dataset. We compare different pretrain settings and evaluate on V2 labels.}
    \label{tab:diving}
    \vspace{-8mm}
\end{table}

\noindent\textbf{Action Recognition.}
We first present action recognition on UCF-101 and HMDB-51 in Table~\ref{tab:recognition}. We report \textit{linear probe} and \textit{finetune} Top-1 accuracy. For a fair comparison, we do not include the works with different evaluation settings and much deeper backbone or non-single modality e.g., optical flow, audio, and text. 

In \textit{linear probe} settings, our method obtains the best result on both UCF-101 and HMDB-51. Even though MLRep~\cite{qian2021enhancing} carefully devises the multi-level feature optimization and temporal modeling, DCLR beats MLRep~\cite{qian2021enhancing} significantly, i.e., 9.1\% and 13\% improvement on UCF-101 and HMDB-51 respectively. 
In addition, DCLR surpasses FAME~\cite{ding2022motion} by 4.2\% on HMDB51, an approach that emphasizes motion pattern learning, demonstrating the effectiveness of our dual contrastive learning framework.

In \textit{finetune} protocol, DCLR still achieves the best results among RGB-only methods. Note that \cite{jenni2020video,huang2021ascnet,chen2021rspnet,zhang2021incomplete} introduce diverse temporal transformations or carefully design temporal pretext tasks. The superiority over them proves our method distills effective spatio-temporal representations by decoupling both levels of data input and feature space.
\begin{table*}[] 
    \small
    \centering
    \begin{tabular}{c|c|cccc|cccc}
        \hline
        \multirow{2}{*}{Method} & \multirow{2}{*}{Backbone} & \multicolumn{4}{c|}{UCF-101} & \multicolumn{4}{c}{HMDB-51} \\
        \cline{3-10}
         & & R@1 & R@5 & R@10 & R@20 & R@1 & R@5 & R@10 & R@20 \\
        \hline
        VCP~\cite{luo2020video} & R3D & 18.6 & 33.6 & 42.5 & 53.3 & 7.6 & 24.4 & 36.3 & 53.6 \\
        Pace~\cite{wang2020self} & R(2+1)D & 25.6 & 42.7 & 51.3 & 61.3 & 12.9 & 31.6 & 43.2 & 58.0 \\
        PRP~\cite{yao2020video} & R(2+1)D & 20.3 & 34.0 & 41.9 & 51.7 & 8.2 & 25.3 & 36.2 & 51.0 \\
        STS~\cite{wang2020statistic} & R3D & 38.3 & 59.9 & 68.9 & 77.2 & 18.0 & 37.2 & 50.7 & 64.8 \\
        VCLR~\cite{kuang2021video} & R2D-50 & 46.8 & 61.8 & 70.4 & 79.0 & 17.6 & 38.6 & 51.1 & \textbf{67.6} \\
        \textbf{DCLR(Ours)} & R(2+1)D & \textbf{54.8} & \textbf{68.3} & \textbf{75.9} & \textbf{82.8} & \textbf{24.1} & \textbf{44.5} & \textbf{53.7} & 64.5 \\
        \hline
    \end{tabular}
    \caption{Results on video retrieval task pretrained on UCF-101. We report R@k (k=1,5,10,20) on UCF-101 and HMDB-51.}
    \label{tab:retrieval}
\end{table*}

Additionally, in Table~\ref{tab:diving}, we report finetune results on a more challenging Diving-48 dataset~\cite{li2018resound}, where all videos share a similar background and only differ in long-term motion patterns. 
In Diving-48, the fine-grained categories are not strongly correlated with static backgrounds anymore. Thus, the result on such a motion-heavy dataset can better display whether the model captures the motion-aware representations.
We compare our method with FAME~\cite{ding2022motion}, which applies motion inductive augmentation to highlight motion patterns in contrastive learning. 
Our method DCLR greatly outperforms FAME with 2$\times$ smaller resolution on both UCF-101 and Kinetics-400 pretrain datasets. 
It demonstrates that DCLR considerably enhances long-range motion pattern modeling and solves the background bias in naive spatio-temporal contrastive learning.

\noindent\textbf{Video Retrieval.}
We show the performance on video retrieval with R@k in Table~\ref{tab:retrieval}. All models are pretrained on UCF-101 for a fair comparison. 
We gain significant improvement on both UCF-101 and HMDB-51 datasets.
It indicates that DCLR formulation encodes static and dynamic characteristics into a more compact manifold.

\subsection{Ablation Study}
\label{ablation}
For further analysis, we dissect our approach on several crucial modules. We pretrain on UCF-101 for 200 epochs, and report the \textit{linear probe} Top-1 accuracy. 

\noindent\textbf{Effectiveness of $\mathcal{L}_{VS}$, $\mathcal{L}_{VD}$, and $\mathcal{L}_{SD}$.}
To analyze the effectiveness of the decoupled learning objective, we make extensive preliminary experiments on various sets of loss. Note that in default settings, for $\mathcal{L}_{VV}$, $\mathcal{L}_{VS}$, and $\mathcal{L}_{VD}$, we use temporally different views of the same video to minimize the loss. Instead, for $\mathcal{L}_{SD}$ that needs to be maximized, we utilize the same view to avoid the collapse issue. 
\begin{table}
    \small
    \centering
    \begin{tabular}{ccccc|c}
         \hline
         $\mathcal{L}_{VV}$  & $\mathcal{L}_{VS}$ & $\mathcal{L}_{VD}$ & $\mathcal{L}_{SD}$ & Same-View & UCF-101 \\
         \hline
         \checkmark & & & & & 48.9 \\
         & \checkmark & & & & 48.1 \\
         & & \checkmark & & & 55.2 \\
         \hline
         & \checkmark & \checkmark & & & 60.2 \\
         & \checkmark & \checkmark & & \checkmark & 41.7 \\
         & \checkmark & \checkmark & \checkmark & & 61.1 \\
         \hline
         \checkmark & \checkmark & & & & 49.3 \\
         \checkmark & & \checkmark & & & 59.4 \\
         \checkmark & \checkmark & \checkmark & & & 61.7 \\
         \checkmark & \checkmark & \checkmark & \checkmark & & 62.4 \\
         \hline
    \end{tabular}
    \caption{Top-1 \textit{linear probe} accuracy on UCF-101. We compare different loss combinations, with $\mathcal{L}_{VV}$, $\mathcal{L}_{VS}$, $\mathcal{L}_{VD}$ to be minimized, $\mathcal{L}_{SD}$ to be maximized. The `same-view' indicates whether adopting the same view samples for $\mathcal{L}_{VS}$ and $\mathcal{L}_{VD}$.}
    \label{tab:mi}
    \vspace{-5mm}
\end{table}
We show the \textit{linear probe} action recognition accuracy on UCF-101 in Table~\ref{tab:mi}. Vanilla contrastive spatio-temporal representation learning baseline is located at the first row in the table, only using $\mathcal{L}_{VV}$. By comparing the baseline and various settings, we reach several important observations. \textbf{First}, considering the first three lines, adoption of $\mathcal{L}_{VS}$ gains a similar result with baseline (0.8\%$\downarrow$) while using $\mathcal{L}_{VD}$ significantly improves the performance (6.3\%$\uparrow$). 
It indicates that naively optimizing $f(v)$ via $\mathcal{L}_{VV}$ nearly equals pulling the video with its static frame, thus losing crucial dynamic motion characters. However, employing $\mathcal{L}_{VD}$ can resist the background shortcuts and boost the representation quality.
The above observation strongly coincides with our motivation about the existing static bias in the contrastive formulation. 
\textbf{Second}, among the middle three lines, we find that jointly minimizing $\mathcal{L}_{VS}$ and $\mathcal{L}_{VD}$ improves the baseline by a large margin (11.3\%$\uparrow$), which means decoupling in data input can validly solve the background bias and help the model capture more compact and comprehensive representations. And if we adopt the same-view setting to formulate $\mathcal{L}_{VS}$ and $\mathcal{L}_{VD}$, the performance will drop dramatically (7.2\%$\downarrow$). It is consistent with our hypothesis that the same view causes the model to deviate from the high-level semantic space and attend to redundant low-level information. Besides, the introduction of the regularization term $\mathcal{L}_{SD}$ further improves the accuracy by 0.9\% compared to only utilization of $\mathcal{L}_{VS}$ and $\mathcal{L}_{VD}$. 
\textbf{Third}, we investigate the combination of $\mathcal{L}_{VV}$ and decoupled contrastive losses in the last four lines. 
Unsurprisingly, integrating $\mathcal{L}_{VV}$ and $\mathcal{L}_{VS}$ seems to bring no improvement (0.4\%$\uparrow$), while jointly optimizing $\mathcal{L}_{VV}$ and $\mathcal{L}_{VD}$ greatly enhances the performance (10.5\%$\uparrow$). 
Also, incorporating $\mathcal{L}_{VV}$ into the combination of $\mathcal{L}_{VS}$ and $\mathcal{L}_{VD}$ only leads to marginal improvement (60.2\% vs 61.7\%).
The phenomenon is reproducible when we add $\mathcal{L}_{SD}$ at the same time (61.1\% vs 62.4\%). 
It is shown that $\mathcal{L}_{VS}$ and $\mathcal{L}_{VD}$ are sufficient for dual contrastive learning. 
Hence, $\mathcal{L}_{VV}$ is not imported in default setting.

\begin{table}
    \small
    \centering
    \begin{tabular}{c|cc}
         \hline
         Setting & UCF-101 & HMDB-51 \\
         \hline
         None & 64.3 & 37.7 \\
         Only $\widetilde{f}(s)$ & 64.5 & 38.3 \\
         Only $\widetilde{f}(d)$ & 67.1 & \textbf{40.1} \\
         Joint $\widetilde{f}(s)$ \& $\widetilde{f}(d)$ & \textbf{67.2} & 39.8 \\
         \hline
    \end{tabular}
    
    \caption{Ablation study on cross-video correspondence search. We compare the baseline and different search settings.}
    \label{tab:cross}
    \vspace{-8mm}
\end{table}

\begin{table}
    \small  
    \centering
    \begin{tabular}{cc|cc}
         \hline
         Num. & Dist. & UCF-101 & HMDB-51 \\
         \hline
         - & - & 64.3 & 37.7 \\
         1 & - & 66.4 & 39.7 \\
         5 & Uniform & 66.9 & 39.5 \\
         5 & Prior & \textbf{67.1} & \textbf{40.1} \\
         10 & Uniform & 65.6 & 38.1 \\
         10 & Prior & 66.3 & 39.8 \\
         \hline
    \end{tabular}
        \caption{Ablation study on cross-video retrieval sample distribution formulation. We compare the results with different numbers and distribution settings.}
    \label{tab:dis}
    \vspace{-8mm}
\end{table}

\noindent\textbf{Cross-video Correspondence Search.}
We first compare different settings for cross-video correspondence search in Table~\ref{tab:cross}. We observe that employing $\widetilde{f}(d)$ to search similar motion patterns considerably facilitates the performance while using $\widetilde{f}(s)$ only spurs marginal improvement. It proves that there exists a severe static bias in the original positive sample construction, i.e., the original positive pairs are well aligned in static scenes but differ in motions. Through our proposed cross-video motion alignment, the static bias issue is solved to some extent. 
It brings nearly 3\% gains on both UCF-101 and HMDB-51 datasets compared to the baseline. 
In default, we discard the search on the static frame $\widetilde{f}(s)$ and only maintain the extractor and queue of frame difference $\widetilde{f}(d)$.

Besides the utilization of cross-video correspondence search, we compare the number of similar pairs and the sample distribution in our corrected motion positive samples. In Table~\ref{tab:dis}, the first line denotes the baseline while the second line means only one sample is taken as positive pair. `Uniform' indicates all pairs are equally important and the final representations are mean averaged. `Prior' means the probability follows pair-wise similarity as Eq.~\ref{prob}. We can see that `Prior' outperforms `Uniform' consistently. 
It validates that the frame difference feature serves as a reliable reference for similar motion pattern retrieval. 
Interestingly, 5 positive samples achieve the highest accuracy while 10 get the worst. 
Theoretically, more abundant positive pairs lead to more accurate estimation, and there should be approximately $2048/101\approx20$ samples per class in the queue. 
However, due to the clip sampling, there are much fewer truly aligned motion patterns as illustrated in Fig.~\ref{fig:teaser}. 
Hence, the performance drops when we retrieve the ten nearest samples to constitute the positive pairs.

\begin{table}
    \small
    \centering
    \begin{tabular}{ccc|cc}
         \hline
         $\mathcal{L}_{ac}$  & $f_s$ & $f_d$ & UCF-101 & HMDB-51 \\
         \hline
         - & - & - & 61.1 & 34.4 \\
         \checkmark & & & 62.7 & 36.1 \\
         \checkmark & \checkmark & & 63.2 & 36.9 \\
         \checkmark & & \checkmark & 63.6 & 37.1 \\
         \checkmark & \checkmark & \checkmark & \textbf{64.3} & \textbf{37.7} \\
         \hline
    \end{tabular}
    \caption{Ablation study on feature activation maps. We compare the results without cross-video retrieval.}
    \label{tab:act}
    \vspace{-8mm}
\end{table}

\noindent\textbf{Feature Activation Maps.}
We detail the effect of using feature activation maps in Table~\ref{tab:act}. The first line denotes only decoupling in data inputs. It is clear that the activation alignment constraint $\mathcal{L}_{ac}$ is effective, and the static-dynamic feature decoupling $f_s$ and $f_d$ further improves performance. 
It is coincident with our motivation that the activation maps provide more concrete spatio-temporal reference to capture both static and dynamic characteristics and mitigate possible static bias.

\begin{figure}
    \centering
    \subfigure[Activation maps $A_s$ of static frame feature.]{
    \includegraphics[width=0.22\linewidth]{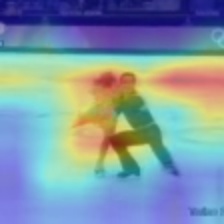}
    \includegraphics[width=0.22\linewidth]{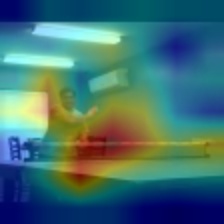}
    \includegraphics[width=0.22\linewidth]{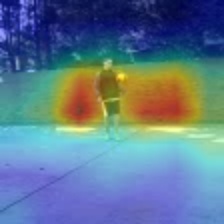}
    \includegraphics[width=0.22\linewidth]{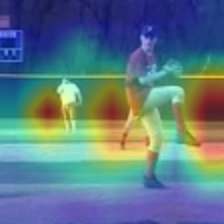}
    }\\
    
    \subfigure[Activation maps $A_d$ of frame difference feature.]{
    \includegraphics[width=0.22\linewidth]{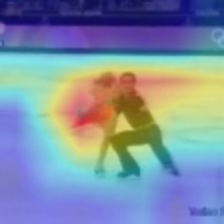}
    \includegraphics[width=0.22\linewidth]{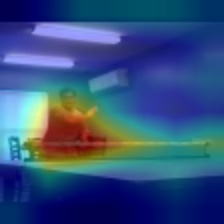}
    \includegraphics[width=0.22\linewidth]{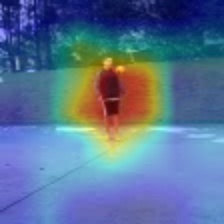}
    \includegraphics[width=0.22\linewidth]{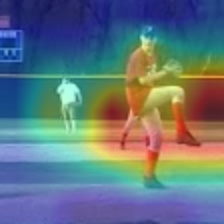}
    }
    
    \caption{Activation maps of static frame and frame difference feature.We observe that $A_s$ and $A_d$ respectively attends to representative backgrounds and dynamic motions.}
    \label{fig:act}
    \vspace{-2em}
\end{figure}

\subsection{Qualitative Analysis}

To better understand how the activation maps work, we visualize the class-agnostic activation maps of the static frame and frame difference features in Fig.~\ref{fig:act}. The static frame features $A_s$ attend to representative background areas like the ping pong table and baseball field. 
In contrast, the frame difference features $A_d$ focus on the moving actors. 
Accordingly, $A_s$ and $A_d$ provide complementary views to understand a video and jointly contributes to unbiased video understanding.

\begin{figure}
    \centering
    \includegraphics[width=\linewidth]{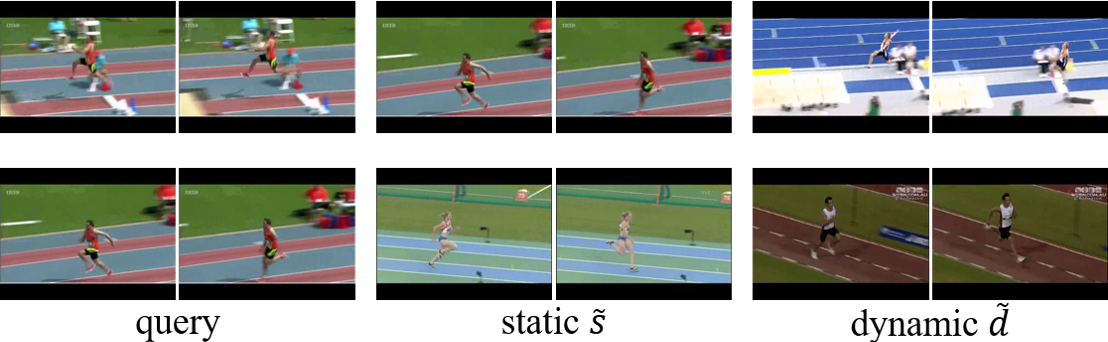}
    \caption{Cross-video search on static and dynamic features. We use two frames to present a clip, and show the Top-1 similar pair for both static scene $s$ and dynamic motion $d$.}
    \label{fig:dist}
\end{figure}

Moreover, we also provide two typical examples of our cross-video retrieval results in Fig.~\ref{fig:dist}. For better illustration, two queries are two views sampled from the same video. 
Through our cross-video search mechanism, we can figure out the exactly aligned clip with the query, especially in dynamic characters. 
For example, we match the query of the first row with another jumping moment as a motion pattern positive pair. And we find another running motion sample across the videos in the second row.
Moreover, the static background of the motion pair appears totally different. 
It verifies that our cross-video search can reduce the background shortcut via the introduction of backgrounds from other videos.
On the contrary, there was little difference between query and static similar pair $\widetilde{s}$ in both motion and background content, which also echos with the quantitative results in Table~\ref{tab:cross} that the dynamic $\widetilde{d}$ makes a great difference but the static $\widetilde{s}$ has a minor effect.

\section{Conclusion}
In this paper, we propose a novel dual contrastive formulation to eliminate the static scene bias in spatio-temporal representation learning. 
We decouple the input RGB video sequence into the static frame and frame difference, then respectively minimize the decoupled loss term to guarantee the comprehensiveness of the RGB feature. 
We further utilize the static and dynamic activation maps as a concrete referrer to filter out redundancy that potentially interferes with learning. 
Through the experiments, we validate the effectiveness of dual contrastive learning formulation that simultaneously encode desired static and dynamic characteristics.

While our work shows some promising results, the state-of-the-art self-supervised performances~\cite{feichtenhofer2021large} of UCF101 and HMDB51 are much higher than ours. But we believe our method can be further boosted through a huger model backbone with greater resolution input. We leave it as the feature work.
\section*{Acknowledgment}
This work was supported in part by the National Natural Science Foundation of China under Grant 61932022, Grant 61971285, Grant 61720106001, and in part by the Program of Shanghai Science and Technology Innovation Project under Grant 20511100100.

\clearpage
\bibliographystyle{ACM-Reference-Format}
\bibliography{ref}


\end{document}